# Optimal Distribution of Solutions for Crowding Distance on Linear Pareto Fronts of Two-Objective Optimization Problems

Hisao Ishibuchi  and  Lie Meng Pang
Guangdong Provincial Key Laboratory of Brain-inspired Intelligent Computation
Department of Computer Science and Engineering, Southern University of Science and Technology, Shenzhen, China
hisao@sustech.edu.cn,  panglm@sustech.edu.cn

*Abstract*—Characteristics of an evolutionary multi-objective optimization (EMO) algorithm can be explained using its best solution set. For example, the best solution set for SMS-EMOA is the same as the optimal distribution of solutions for hypervolume maximization. For NSGA-III, if the Pareto front has intersection points with all reference lines, all of those intersection points are the best solution set. For MOEA/D, the best solution set is the set of the optimal solution of each sub-problem. Whereas these EMO algorithms can be analyzed in this manner, the best solution set for the most well-known and frequently-used EMO algorithm NSGA-II has not been discussed in the literature. This is because NSGA-II is not based on any clear criterion to be optimized (e.g., hypervolume maximization, distance minimization to the nearest reference line). As the first step toward the best solution set analysis for NSGA-II, we discuss the optimal distribution of solutions for the crowding distance under the simplest setting: the maximization of the minimum crowding distance on linear Pareto fronts of two-objective optimization problems. That is, we discuss the optimal distribution of solutions on a straight line. Our theoretical analysis shows that the uniformly distributed solutions are not the best solution set. However, it is also shown by computational experiments that the uniformly distributed solutions (except for the duplicated two extreme solutions at each edge of the Pareto front) are obtained from a modified NSGA-II with the ($\mu$ + 1) generation update scheme.

## I. Introduction

When we analyze an existing evolutionary multi-objective optimization (EMO) algorithm, its important characteristic is the best solution set for the algorithm. This is because the best solution set or a similar solution set is obtained after enough generations if the search is not trapped in local optima. For the same reason, the best solution set is important when we design a new EMO algorithm. In general, each EMO algorithm has a different best solution set. This is totally different from the single-objective optimization field where the best solution is the same for all algorithms, which is the optimal solution of the given single-objective optimization problem. In the EMO field, the goal is to find a well-distributed solution set over the entire Pareto front. However, we do not have a consensus about the quantitative definition (e.g., maximization of the hypervolume indicator value, maximization of the uniformity indicator values) about the best distribution of solutions in such a target solution set. As a result, various EMO algorithms have been proposed in the literature. The concept of the best solution set is much more important when some preference information is available from the decision maker. In this case, the best solution set needs to be consistent with the decision maker's preference.

The best solution set for an EMO algorithm is usually a set of Pareto optimal solutions. This is because there exists a better solution (i.e., dominating solution) for each dominated solution. Pareto optimal solutions have no such a better solution. One clear feature of each EMO algorithm can be explained by its own mechanism to choose the best solution set among Pareto optimal solutions. For example, the best solution set for a hypervolume-based algorithm (e.g., SMS-EMOA[1], [2]) is the optimal distribution $\mu$ solutions for hypervolume maximization where $\mu$ is the population size. This is because such an EMO algorithm maximizes the hypervolume indicator value of the current population. Since the late 2000s [3], [4], the optimal distribution of solutions for hypervolume maximization has been actively studied in the EMO community [5]-[10]. Those studies have shown the following: (i) the optimal distribution is uniform on linear Pareto fronts of two-objective problems, (ii) the optimal distribution is not uniform on non-linear Pareto fronts, and (iii) the optimal distribution depends on the location of a reference point for hypervolume calculation. The optimal distribution of solutions for other indicators such as IGD (Inverted Generational Distance) and IGD+ have also been studied (e.g., [11]-[13]).

For a decomposition-based algorithm (e.g., MOEA/D [14], NSGA-III [15]), the best solution set is usually the same as or similar to the set of all intersection points of the reference lines (search directions) with the Pareto front. NSGA-III selects the nearest non-dominated solution to each reference line. Thus, if all reference lines have intersection points with the Pareto front, the best solution set is the set of all of those intersection points. MOEA/D searches for the best solution for each subproblem. Thus the optimal solution set is the set of the optimal solution of each subproblem. The best solution for each subproblem is

This work was supported by National Natural Science Foundation of China (Grant No. 62376115, 62250710682), Guangdong Provincial Key Laboratory (Grant No. 2020B121201001).
Corresponding Author: Hisao Ishibuchi

usually on the weight vector (i.e., search direction) or near the weight vector if the weight vector has an intersection point with the Pareto front. As a result, the best solution set for MOEA/D is the same as or similar to the set of all intersection points when all weight vectors have intersection points.

For indicator-based and decomposition-based algorithms, it is easy to discuss the best solution set for each algorithm. However, for NSGA-II (Nondominated Sorting Genetic Algorithm II) [16], it is not easy to discuss the best solution set because it does not directly optimize any criterion. In Fig. 1, we show the final populations of NSGA-II, MOEA/D-PBI with the penalty parameter value 5, and SMS-EMO on the three-objective DTLZ1 [17] under the standard settings using PlatEMO [18]. In Fig. 1, the final population of NSGA-II shows no clear pattern (i.e., regularity). Experimental results on the three-objective DTLZ2 [17] and WFG4 [19] are shown in Fig. 2 and Fig. 3, respectively. Fig. 2 shows that uniformly distributed solutions are not the best solution set for SMS-EMOA. Fig. 3 shows the necessity of normalization in MOEA/D for WFG4 since it has differently scaled objectives (e.g., the $f_3$-axis is three times larger than the $f_1$-axis in Fig. 3).

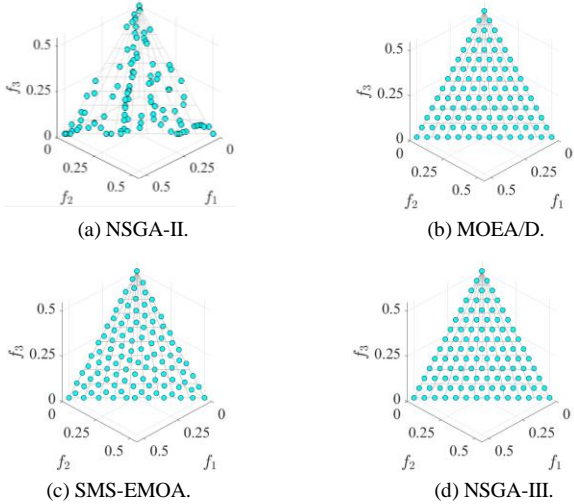

Fig. 1. A final population of each algorithm on the 3-objective DTLZ1.

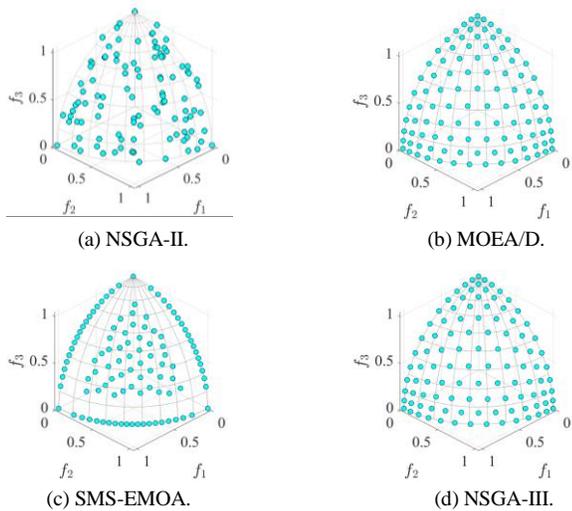

Fig. 2. A final population of each algorithm on the 3-objective DTLZ2.

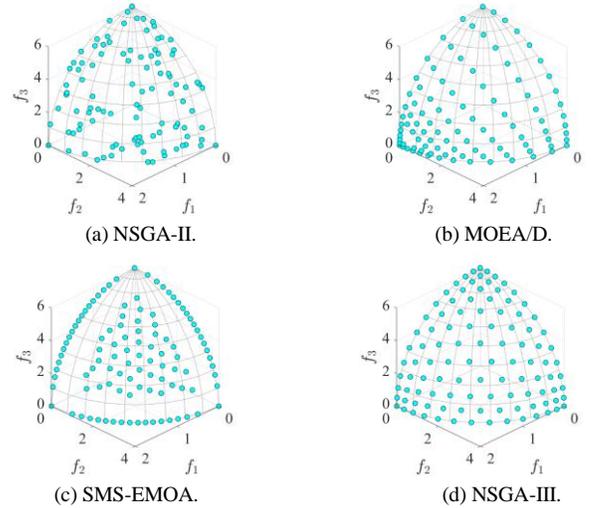

Fig. 3. A final population of each algorithm on the 3-objective WFG4.

In Figs. 1-3, the final populations of MOEA/D, SMS-EMOA and NSGA-III are good approximations of the best solution sets. For MOEA/D and NSGA-III, the intersection points with the Pareto front are obtained. For SMS-EMOA, the final population in each figure is an approximately optimal solution set for hypervolume maximization, which is uniform on the linear Pareto front in Fig. 1, and non-uniform on the non-linear Pareto fronts in Fig. 2 and Fig. 3.

However, for NSGA-II, we cannot observe any clear patterns (i.e., regularities) in the final populations in Figs. 1-3 (a) where the solution distributions look random. The reason for such a random distribution can be explained as follows. The crowding distance of NSGA-II is calculated on each axis of the objective space. Thus, except for the case of two objectives, the crowding distance cannot evaluate the distance to the neighboring solutions (i.e., local density) in the objective space [25]. As a result, we cannot obtain well-distributed solutions by NSGA-II for the three-objective problems. One question is whether the final populations in Figs. 1-3 (a) are the best solution sets for NSGA-II. This is a very difficult question for three-objective problems. For example, the optimal distribution of solutions for hypervolume maximization has been clearly shown only for two-objective problems. So, as the first step toward the best solution set analysis for NSGA-II, we focus on linear Pareto fronts of two-objective problems in this paper. That is, we discuss the optimal distribution of solutions on a straight line. Even for such a simple case, we obtain the following counterintuitive observation: the optimal distribution for the maximization of the minimum crowding distance is not uniform. In the optimal distribution, two solutions are always overlapping. For example, in the optimal distribution of eight solutions, they are on four uniformly-distributed locations, and two solutions are overlapping on each location. However, such a counterintuitive best distribution is not obtained by NSGA-II. Somewhat randomized distributions are obtained by NSGA-II. This is because many solutions with the smallest crowding distance values are removed in NSGA-II without crowding distance recalculation. If all solutions are non-dominated in the merged population of the current and offspring populations, μ solutions with the smallest crowding distance values are removed with no crowding distance recalculation. This is explained in Fig. 4 (a) for the case of $\mu = 5$ where the five gray solutions are removed. To further analyze the search behavior of NSGA-II, we implement a modified NSGA-II algorithm with the $(\mu + 1)$ generation update scheme where only a single solution is generated at each generation. That is, only one solution is removed from the merged population. This is explained in Fig. 4 (b) for the case of $\mu = 5$ where the gray solution is removed. Using this variant of NSGA-II, we demonstrate that uniform distributions are obtained. However, even in this case, two solutions are overlapping on the two edges of the linear Pareto front. The reasons for these observations are explained later in this paper.

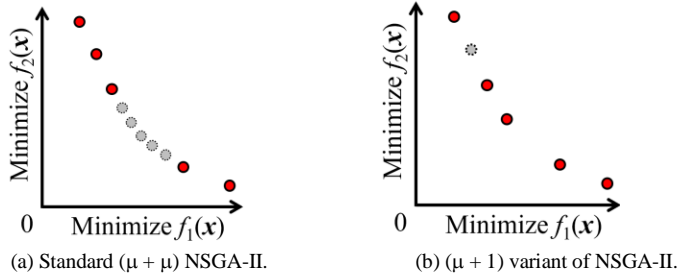

(a) Standard $(\mu + \mu)$ NSGA-II.  (b) $(\mu + 1)$ variant of NSGA-II.

Fig. 4. Comparison of the standard $(\mu + \mu)$ NSGA-II algorithm and the modified $(\mu + 1)$ NSGA-II algorityhm.

The main contribution of this paper is to show the following: (i) the optimal distribution of solutions to maximize the minimum crowding distance is not uniform, and (ii) unform distributions are obtained by the $(\mu + 1)$ variant of NSGA-II.

This paper is organized as follows. In Section II, NSGA-II is briefly explained. In Section III, the optimal distributions of a small number of solutions are theoretically shown for the maximization of the minimum crowding distance on a linear Pareto front. The obtained results can be easily extended to the case of many solutions. In Section IV, we report experimental results using two versions of NSGA-II. One is the standard NSGA-II with the $(\mu + \mu)$ generation update scheme, and the other is a NSGA-II variant with the $(\mu + 1)$ generation update scheme. Finally, we conclude this paper in Section V.

## II. NSGA-II ALGORITHM

In the EMO field, NSGA-II [16] is the most well-known and frequently-used algorithm. Even now, NSGA-II is used in many application papers whereas it was proposed more than 20 years ago. As shown in Figs. 1-3, NSGA-II cannot find well-distributed solutions for three-objective optimization problems. Moreover, its convergence ability is often severely deteriorated by the increase in the number of objectives as demonstrated using the DTLZ test problems [17] in the late 2000s [20], [21]. However, despite of these well-known weak points, recent studies [22]-[24] showed that NSGA-II usually works well on real-world applications and often outperforms state-of-the-art algorithms in many cases whereas it is clearly outperformed by other algorithms on artificial test problems (e.g., DTLZ [17]).

Here, we briefly explain the generation update mechanism of NSGA-II. Let $N$ be the population size. From the current population with $N$ solutions, $N$ offspring are generated. Then, from the merged population with $2N$ solutions, $N$ solutions are selected for the next generation. That is, NSGA-II is based on the $(\mu + \mu)$ scheme. Each solution in the merged population is evaluated by the non-dominated sorting. First, non-dominated solutions in the merged population are removed and included in the first front. Next, non-dominated solutions among the remaining solutions are removed and included in the second front. In this manner, all solutions are sorted. The sorting result is the primary fitness of each solution. All solutions in the first front have the best fitness, and all solutions in the second front have the second best fitness. To differentiate solutions in the same front, the crowding distance of each solution is calculated in each front as explained in Fig. 5. The crowding distance of a solution is the sum of the distance between its two adjacent solutions on each axis of the objective space. For example, the crowding distance of solution C in Fig. 5 is calculated as $c_1 + c_2$. Two extreme solutions (i.e., solutions A and F in Fig. 5) with the maximum and minimum values of each objective have an infinitely large value as the crowding distance.

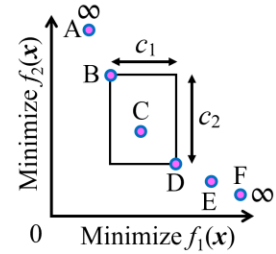

Fig. 5. Crowding distance calculation.

The next population is selected based on the primary fitness (i.e., non-dominated sorting) and the secondary fitness among the solutions in the same front (i.e., crowding distance: the larger, the better). Except for early generations, the number of non-dominated solutions in the merged population is usually larger than the population size. Thus, the next population is usually selected from the first front using the crowding distance. So, in this paper, we discuss the optimal distribution of solutions for the crowding distance.

## III. OPTIMAL DISTRIBUTION FOR CROWDING DISTANCE

In NSGA-II, Pareto optimal solutions are always included in the first front. Non-Pareto optimal solutions can be included in the first or lower front. If solution A is not Pareto optimal, there exists at least one solution which dominates solution A. As a result, solution A can be included in the second front. Since the next population is usually selected from the first front, we can assume that the best solution set for NSGA-II is a subset of the Pareto optimal solution set. Then, the question is which is the best subset for NSGA-II. Since solutions with large crowding distance are selected for the next population, the best subset is a set of Pareto optimal solutions with large crowding distance. Thus, in this paper, we assume that the best solution set for NSGA-II is a subset of Pareto optimal solutions where the minimum crowding distance is maximized. Since it is very difficult to discuss the optimal distribution of solutions for the case of three or more objectives (see Figs. 1-3 (a)), in this paper, we focus on the optimal distribution on linear Pareto fronts of two-objective problems as the first step toward the optimal solution set analysis for NSGA-II.

*A. Assumptions in This Paper*

We assume that the Pareto front of a two-objective problem is a line between (1, 0) and (0, 1) in the normalized objective space as shown in Fig. 6 (a). We also assume that the two extreme points (1, 0) and (0, 1) of the Pareto front are included in the best solution set for NSGA-II and they have an infinitely large crowding distance as shown in Fig. 6 (b). Thus, our task in this paper is to find the best location of each of the other solutions for maximizing the minimum crowding distance.

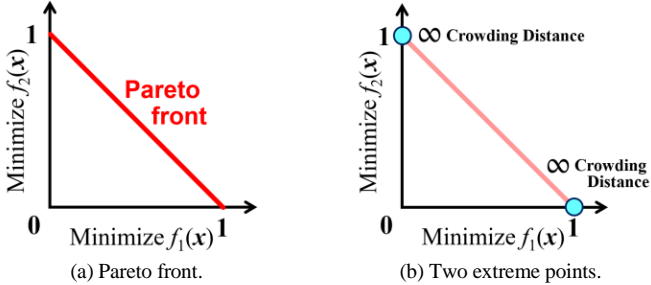

(a) Pareto front.  (b) Two extreme points.
Fig. 6. The linear Pareto front and the two extreme points.

*B. Optimal Distributions of Solutions*

**Optimal Distribution of Three Solutions:** As the simplest case, let us discuss the optimal distribution of three solutions for maximizing the minimum crowding distance. Since we assume that we have two extreme solutions at (0, 1) and (1, 0), our task is to find the best location of the other solution. Since we also assume that the best subset for NSGA-II is a subset of Pareto optimal solutions, the location of the other solution can be denoted as $(x, 1-x)$ on the Pareto front, i.e., solution A in Fig. 7 (a). It is clear that the crowding distance of solution A is 2 (i.e., $x + (1-x) + (1-x) + x$) independent of the value of $x$ in its location $(x, 1-x)$. Thus, any location of solution A is optimal as far as solution A is on the Pareto front as shown in Fig. 7 (b). This result is somewhat counterintuitive since we implicitly think that the distribution is optimal when solution A is at the center of the Pareto front: (0.5, 0.5).

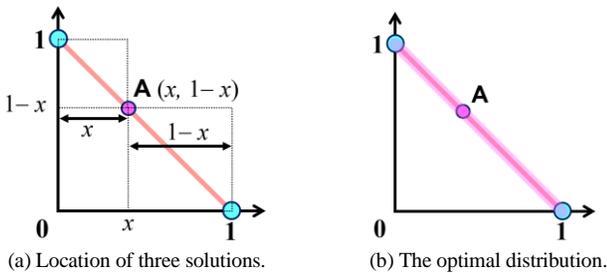

(a) Location of three solutions.  (b) The optimal distribution.
Fig. 7. The optimal distribution of three solutions.

**Optimal Distribution of Four Solutions:** In this case, from the symmetric nature of the problem, we denote two solutions A and B as $(x, 1-x)$ and $(1-x, x)$ as shown in Fig. 8 (a). From this figure, we can see that the crowding distance of solutions A and B is calculated as $2(1-x)$. It is clear that this crowding distance is maximized when $x = 0$. That is, A is on one extreme solution, and B is on the other extreme solution as shown in Fig. 8 (b). Thus, the optimal distribution of four solutions are two overlapping solutions on the two extreme points. In this distribution, the minimum crowding distance is 2. When the four solutions are uniformly distributed (i.e., $x = 1/3$ in Fig. 8 (a)), the minimum crowding distance is $2(1-x) = 4/3$. The optimal distribution of four solutions in Fig. 8 (b) is also counterintuitive whereas solution A and B have the largest crowding distance (which is clearly larger than the case of the uniform distribution: 2 vs. 4/3).

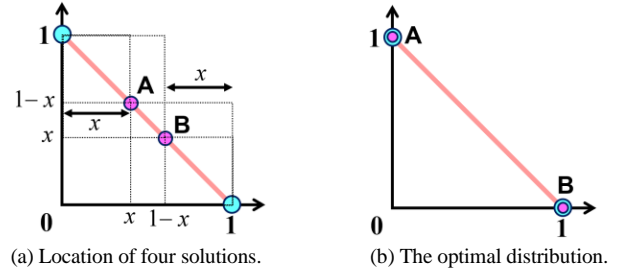

(a) Location of four solutions.  (b) The optimal distribution.
Fig. 8. The optimal distribution of four solutions.

**Optimal Distribution of Five Solutions:** In this case, we insert one solution C between solutions A and B in Fig. 8 (a). The location of C is denoted as $(y, 1-y)$ as shown in Fig. 9 (a). The crowding distance of the three solutions A, B and C in Fig. 9 (a) is calculated as $2y$, $2(1-y)$, and $2(1-2x)$, respectively. To maximize the minimum crowding distance (i.e., maximize $\min\{2y, 2(1-y), 2(1-2x)\}$), $y$ should be at $y = 0.5$ and $x$ should be in $[0, 1/4]$. This means that point C is the center of the Pareto front, and points A and B are near the two extreme solutions as explained in Fig. 9 (b). In this case, the minimum crowding distance is 1, which is the largest minimum crowding distance. Only when $x = 1/4$, we have the uniform distribution of five solutions. If we allow non-symmetric distribution of five solutions, the optimality condition of solutions A and B for the maximization of the minimum crowding distance is as follows: Solutions A and B on the Pareto front have at least the Manhattan distance of 1 so that the crowding distance of solution C is equal to or larger than 1. Thus, the solution sets in Fig. 10 (a) and (b) are also the optimal distribution of solutions.

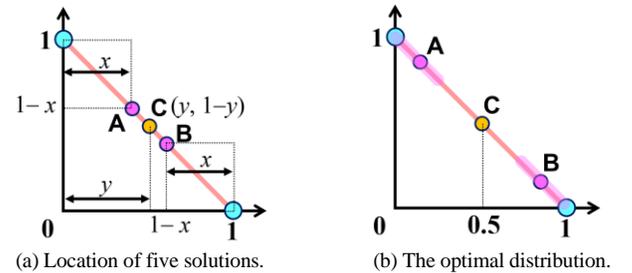

(a) Location of five solutions.  (b) The optimal distribution.
Fig. 9. The optimal distribution of five solutions.

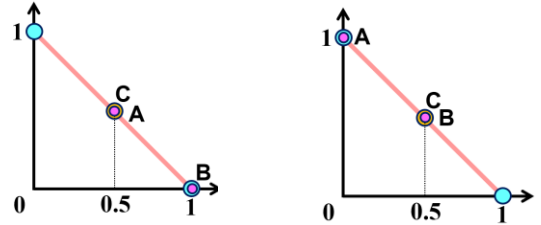

Fig. 10. Examples of other optimal distributions of five solutions.

**Optimal Distribution of Six Solutions:** Based on the symmetric nature of the problem, we specify four solutions A, B, C, and D as shown in Fig. 11 (a). The crowding distance of each solution is as follows: A and B: $2(1 - x - y)$, and C and D: $2x$. To maximizing the minimum crowding distance, we have $x = 0.5$ and $y = 0$. This means that C is at the top-left extreme point (0, 1), A and B are at the center (0.5, 0.5) of the Pareto front, and D is at the bottom-right extreme point (1, 0). This optimal distribution is shown in Fig. 11 (b). In the optimal distribution of six solutions in Fig. 11 (b), two solutions are overlapping at each of the three locations: the two extreme points and the center of the Pareto front. In this case, the minimum crowding distance is 1. If the six solutions are uniformly distributed (i.e., $x = 0.4$, $y = 0.2$), the minimum crowding distance is 0.8. Again, the optimal distribution in Fig. 11 (b) is counterintuitive whereas it maximizes the minimum crowding distance.

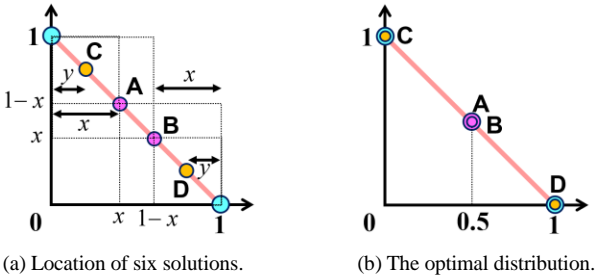

(a) Location of six solutions.  (b) The optimal distribution.

Fig. 11. The optimal distribution of six solutions.

**Optimal Distributions of Seven and Eight Solutions:** In the same manner, we can find the optimal distribution of seven or more solutions. In Fig. 12, we show the optimal distributions of seven and eight solutions. In the case of seven solutions, Fig. 12 (a) is not the unique optimal distribution. For example, solution E shown by the yellow circle in Fig. 12 (a) can be on solution D. Any assignments of seven solutions to the four locations in Fig. 12 (a) with the maximum of two solutions at each location are optimal. Thus, we have four different optimal distributions of seven solutions. For the case of eight solutions, Fig. 12 (b) is the unique optimal distribution.

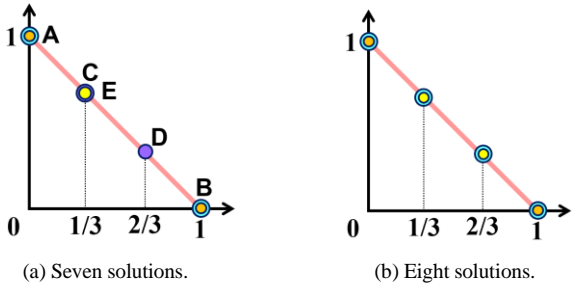

(a) Seven solutions.  (b) Eight solutions.

Fig. 12. The optimal distributions of seven and eight solutions.

From the theoretical analysis in this section, we can see that two solutions are overlapping in the optimal distribution of solutions on the linear Pareto front for the maximization of the minimum crowding distance. This suggests that NSGA-II does not maximize the minimum crowding distance whereas it removes solutions with the smallest crowding distance values from the merged population. In the next section, we discuss this issue through computational experiments.

## IV. COMPUTATIONAL EXPERIMENTS

### A. Results of the Standard NSGA-II Algorithm

In this subsection, we apply the standard NSGA-II with the $(\mu + \mu)$ generation update scheme to the two-objective DTLZ1 test problem with a linear Pareto front. The number of decision variables is specified as six (i.e., $n = 6$). We examine various specifications of the population size. NSGA-II is applied to DTLZ1 under the following setting:

Population size: $N$ ($N$ = 3, 4, 5, ...).
Termination condition: 10,000 generations.
Crossover: SBX crossover with distribution index 20.
Crossover probability: 1.
Mutation: Polynomial mutation with distribution index 20.
Mutation probability: $1/n$.
Number of independent runs: 10.
Experiment Platform: PlatEMO [18].

After preliminary computational experiments, we obtained the following observations about the crowding distance:

**Implementation Dependent Observation**: Two solutions in the final population are overlapping at each edge of the Pareto front. Those two solutions (i.e., four solutions in total) have an infinitely large crowding distance.

This observation can be explained as follows. Let us assume that we have five solutions on the linear Pareto front in the previous section (i.e., the straight line between (0, 1) and (1, 0)) as follows: A (0, 1), B (0, 1), C (0.5, 0.5), D (1, 0), E (1, 0). Based on the first objective values, these solutions are sorted in an ascending order as ABCDE. Thus, an infinitely large crowding distance is assigned to A and E. These solutions are also sorted in an ascending order of the second objective values. If the order is exactly the reverse order of the ascending order for the first objective values (i.e., EDCBA), the same two solutions A and E have an infinitely large crowding distance. However, since A and B (and D and E) are on the same point on the Pareto front, the ascending order of the second objective values can be DECAB. In this case, solutions B and D in addition to solutions A and E have an infinitely large crowding distance. Once those four solutions are obtained at the two edges of the Pareto front, it is likely that those solutions stay in the population forever since they are always in the first front and have the best crowding distance. This is the reason why the final population of NSGA-II often includes two overlapping extreme solutions at each edge of the Pareto front. Since two solutions at each edge have an infinitely large crowding distance, our theoretical analysis in the previous section for $N$ solutions corresponds to our experimental results with $(N + 2)$ solutions in this section.

Among ten runs of NSGA-II, we show the final populations of six runs due to the paper length limit. Fig. 13 shows the results with the population size 5. Two solutions are at each edge of the Pareto front, and the location of the other solution looks random in Fig. 13. This observation is consistent with our analysis in Fig. 7 for three solutions. Fig. 14 shows the results with the population size 6. The locations of the two solutions look random. This observation is not consistent with our theoretical analysis in Fig. 8 for four solutions where all solutions are on the two edges of the linear Pareto front.

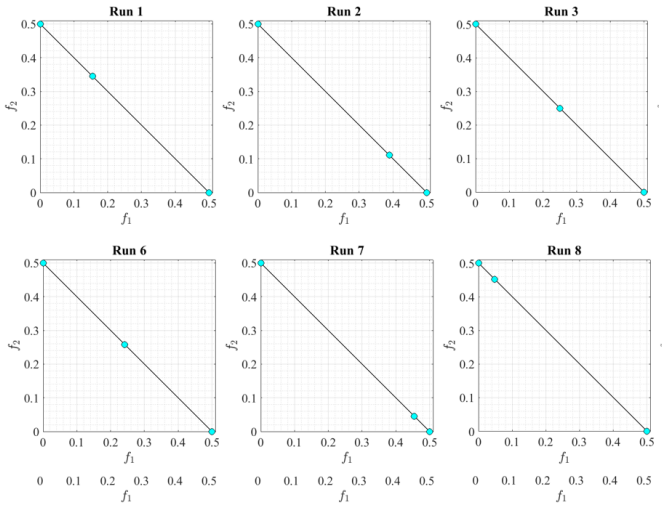

Fig. 13. Experimental results by NSGA-II (5 solutions).

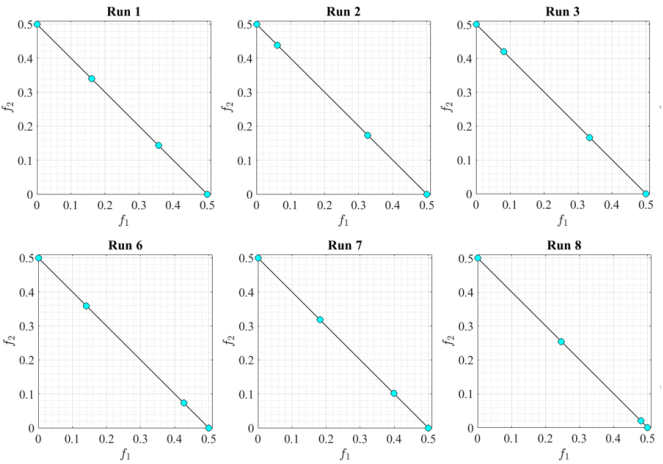

Fig. 14. Experimental results by NSGA-II (6 solutions).

Experimental results for population size 7, 8, 20 and 50 are shown in Figs. 15-18. As in Figs. 13-14, we cannot observe any clear patterns (i.e., regularities) in Figs. 15-18. Solution distributions in Figs. 15-18 have some randomness.

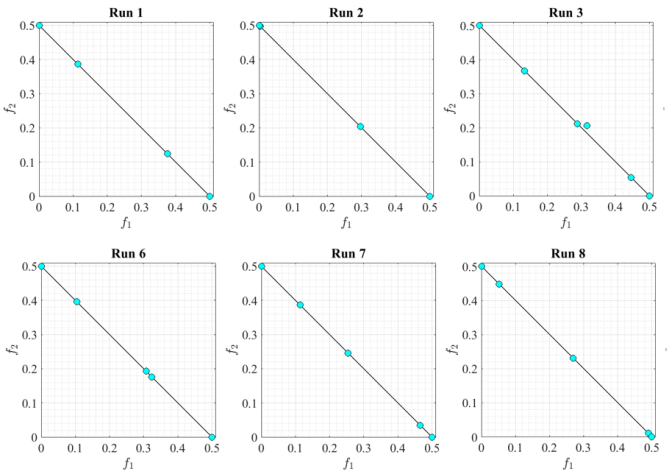

Fig. 15. Experimental results by NSGA-II (7 solutions).

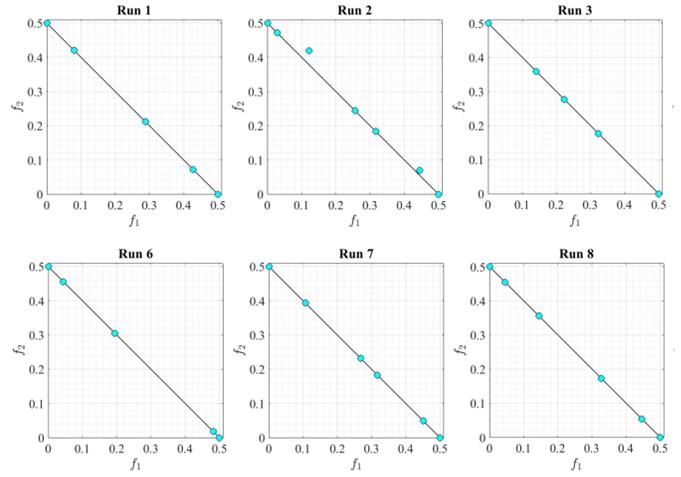

Fig. 16. Experimental results by NSGA-II (8 solutions).

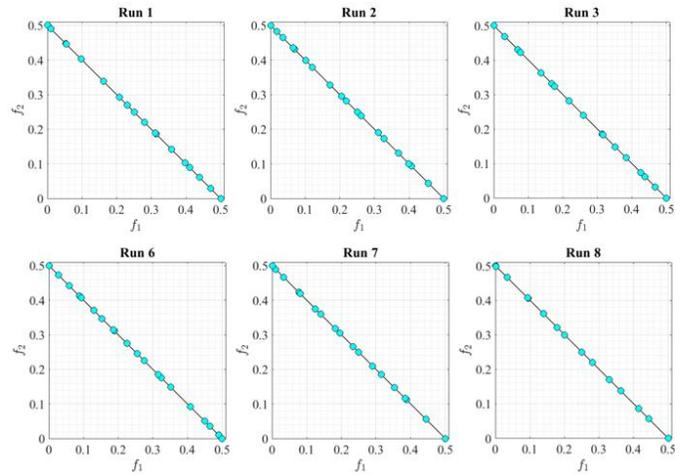

Fig. 17. Experimental results by NSGA-II (20 solutions).

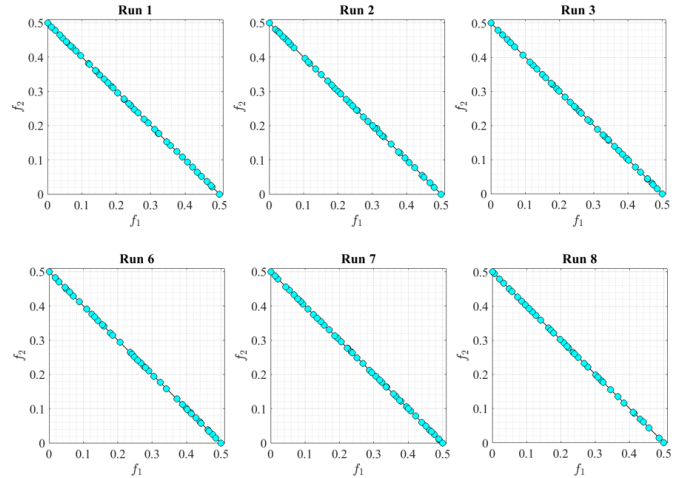

Fig. 18. Experimental results by NSGA-II (50 solutions).

## B. Result of the Steady State NSGA-II Algorithm

Experimental results by the modified NSGA-II with the (μ + 1) generation update scheme are shown in Figs. 19-24.

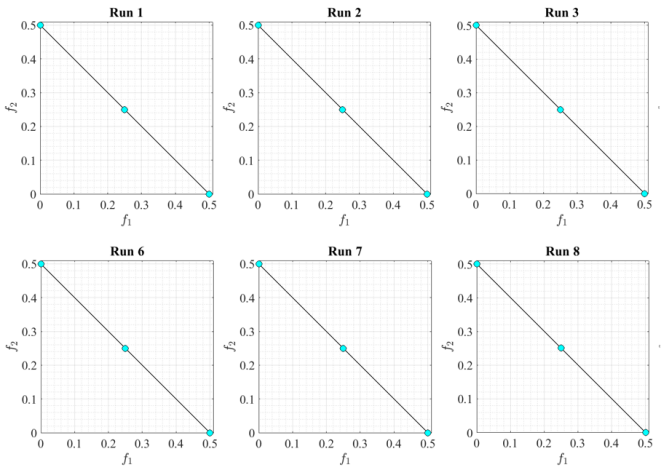

Fig. 19. Experimental results by (μ+1) NSGA-II (5 solutions).

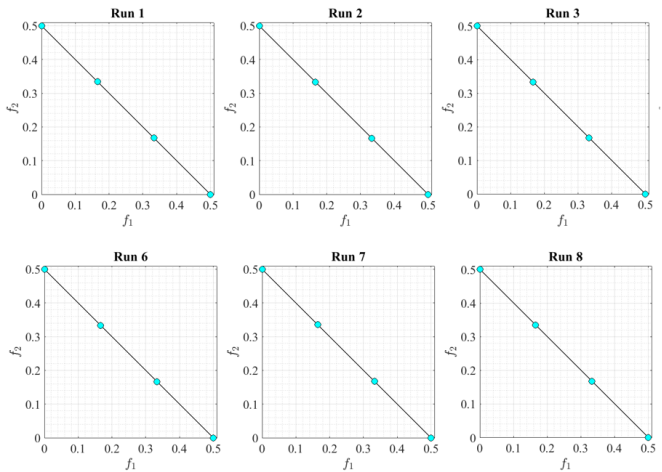

Fig. 20. Experimental results by (μ+1) NSGA-II (6 solutions).

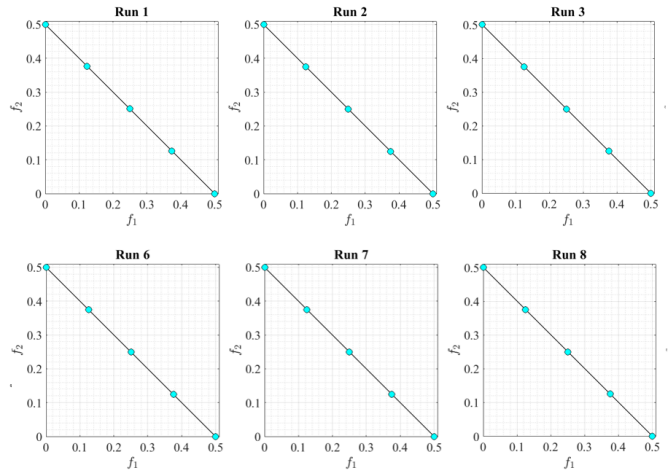

Fig. 21. Experimental results by (μ+1) NSGA-II (7 solutions).

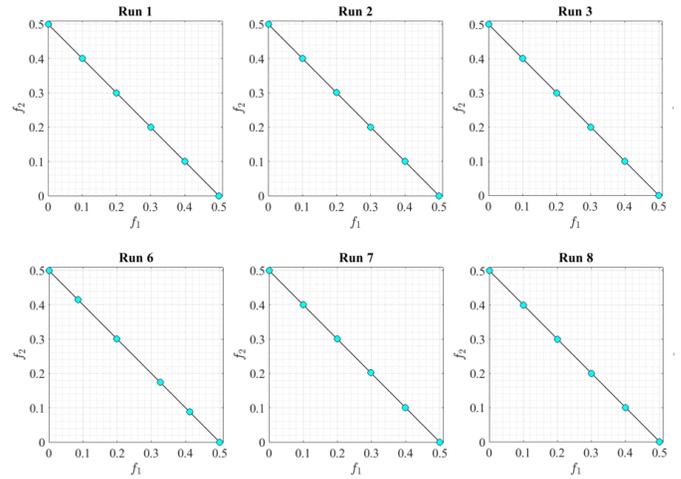

Fig. 22. Experimental results by (μ+1) NSGA-II (8 solutions).

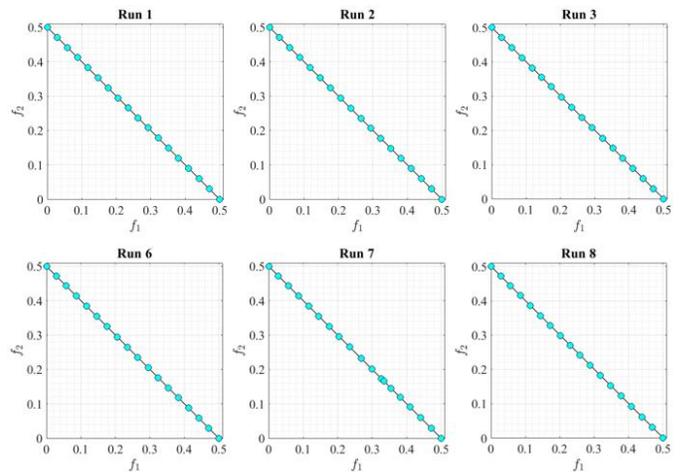

Fig. 23. Experimental results by (μ+1) NSGA-II (20 solutions).

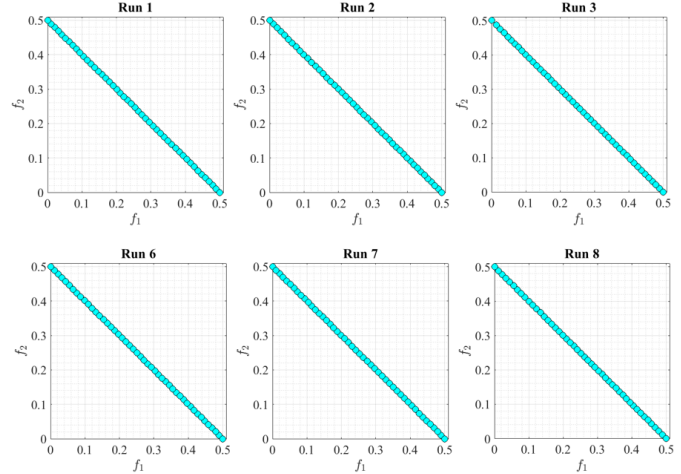

Fig. 24. Experimental results by (μ+1) NSGA-II (50 solutions).

One clear observation from the experimental results by the modified (μ + 1) NSGA-II algorithm in this subsection is that the solutions are uniformly distributed on the linear Pareto front (except for the two overlapping solutions at each edge).

In the modified (μ + 1) NSGA-II algorithm, one offspring is added to the current population. If all solutions are non-dominated, one solution with the smallest crowding distance is removed. Thus, it seems that the crowding distance of each solution is maximized. However, as we have explained, the optimal distribution for the maximization of the minimum crowding distance is not uniform. Experimental results in this subsection shows that NSGA-II maximizes the uniformity of solutions instead of the crowding distance of each solution. This issue is further discussed below.

For example, in SMS-EMOA, the worst solution with the minimum hypervolume contribution is removed in the $(\mu + 1)$ generation update scheme to maximize the hypervolume of the population. In this generation update mechanism, the point is that the contribution of each solution is calculated. Thus, if we want to maximize the minimum crowding distance in NSGA-II, we need to calculate the contribution of each solution to the minimum crowding distance (instead of the crowding distance itself). For clarifying this discussion, let us assume that we have five solutions in the merged population in Fig. 25 (a). In this case, the $(\mu + 1)$ generation update scheme based on the crowding distance in NSGA-II removes solution B since it has the smallest crowding distance. After removing B, we have the four solutions in Fig. 25 (b). However, if we remove solution C as in Fig. 25 (c), the better solution set with a larger minimum crowding distance is obtained. That is, solution C is the worst solution in terms of the contribution to the minimum crowding distance in Fig. 25 (a). This discussion clearly explains that the generation update scheme of NSGA-II does not maximize the minimum crowding distance.

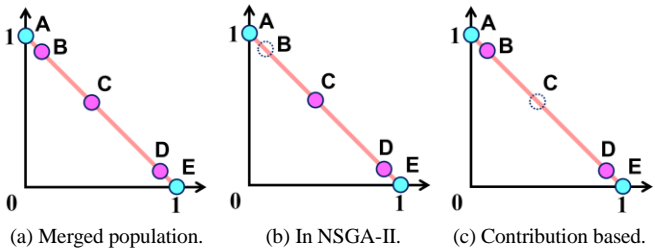

(a) Merged population.   (b) In NSGA-II.   (c) Contribution based.

Fig. 25. Removal of a single solution from the five solutions in (a). The worst crowding distance removal in (b), and the worst contribution removal in (c).

## V. CONCLUDING REMARKS AND FUTURE RESEARCH TOPICS

In this paper, first, we theoretically showed that the optimal distribution of solutions for the maximization of the minimum crowding distance on the linear Pareto front of a two-objective problem is not uniform. In the optimal distribution of solutions, two solutions are always overlapping on each of uniformly distributed locations on the linear Pareto front. Next, we compared the theoretically obtained optimal distributions with the final populations of NSGA-II. However, we could not find any clear patterns (i.e., regularities) in the final populations of NSGA-II. Then, we used a steady state variant of NSGA-II with the $(\mu + 1)$ generation update scheme. Using this variant, we demonstrated that the solutions in the final populations are always uniformly distributed. This observation together with our theoretical analysis about the optimal distribution showed that NSGA-II does not maximize the minimum crowding distance. Based on these results, we explained why the crowding distance-based generation update mechanism of NSGA-II does not maximize the minimum crowding distance whereas it removed the solution with the minimum crowding distance by its generation update mechanism.

One interesting future research topic is to examine why the crowding distance-based generation update mechanism in NSGA-II maximizes the uniformity of solutions instead of the minimum crowding distance. It is also an interesting future research topic to examine the best solution sets for other well-known EMO algorithms. In this paper, we examined the best solution sets for NSGA-II on the linear Pareto front in the two-dimensional objective space. In this case, the uniformly distributed solution sets are obtained by the modified $(\mu + 1)$ NSGA-II as shown in Figs. 19-24. It is also an interesting future research topic to examine the best solution sets in the case of three or more objectives. As we explained in Figs. 1-3 (a), we could not find any clear patterns (i.e., regularities) in the final populations of the standard $(\mu + \mu)$ NSGA-II for three-objective problems. For comparison, in Fig. 26, we show the final populations of the modified $(\mu + 1)$ NSGA-II on the three-objective DTLZ1 and DTLZ2 problems. We cannot observe any clear uniformity improvement of solutions in Fig. 26 with $(\mu + \mu)$ from Fig. 1 (a) and Fig. 2 (a) with $(\mu + 1)$. We also calculate the crowding distance value of each solution in the final population in Fig. 1 (a) by the standard $(\mu + \mu)$ NSGA-II and the final population in Fig. 26 (a) by the modified $(\mu + 1)$ NSGA-II. The histogram of the crowding distance values in each final population is shown in Fig. 27. We can observe in Fig. 27 that Fig. 27 (b) has a larger minimum crowding distance than Fig. 27 (a). This means that the final population of the modified $(\mu + 1)$ NSGA-II in Fig. 26 (a) has a larger minimum crowding distance than the final population of the standard $(\mu + \mu)$ NSGA-II in Fig. 1 (a). However, Fig. 27 (b) still has a large diversity in the crowding distance values in the final population of the modified $(\mu + 1)$ NSGA-II. This may mean that the minimum crowding distance is not maximized in Fig. 27 (b). These observations will be useful in analyzing the search behavior of NSGA-II in future studies.

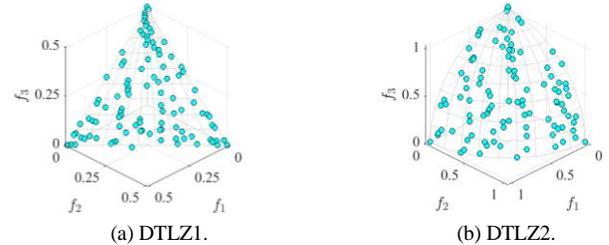

(a) DTLZ1.   (b) DTLZ2.

Fig. 26. Final populations of the $(\mu + 1)$ NSGA-II algorithm on the 3-objective DTLZ1 and DTLZ2 problems.

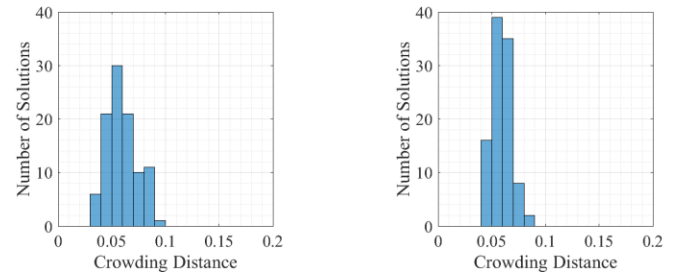

(a) $(\mu+\mu)$ NSGA-II in Fig. 1 (a).   (b) $(\mu+1)$ NSGA-II in Fig. 26 (a).

Fig. 27. Distribution of crowding distance values in the final population in Fig. 1 (a) and Fig. 26 (a) on the three-objective DTLZ1 problem.